# Dempster–Shafer clustering using Potts spin mean field theory

**M. Bengtsson, J. Schubert**

**Abstract** In this article we investigate a problem within Dempster–Shafer theory where $2^q - 1$ pieces of evidence are clustered into $q$ clusters by minimizing a metaconflict function, or equivalently, by minimizing the sum of weight of conflict over all clusters. Previously one of us developed a method based on a Hopfield and Tank model. However, for very large problems we need a method with lower computational complexity. We demonstrate that the weight of conflict of evidence can, as an approximation, be linearized and mapped to an antiferromagnetic Potts spin model. This facilitates efficient numerical solution, even for large problem sizes. Optimal or nearly optimal solutions are found for Dempster–Shafer clustering benchmark tests with a time complexity of approximately $O(N^2 \log^2 N)$. Furthermore, an isomorphism between the antiferromagnetic Potts spin model and a graph optimization problem is shown. The graph model has dynamic variables living on the links, which have a priori probabilities that are directly related to the pairwise conflict between pieces of evidence. Hence, the relations between three different models are shown.

## 1
## Introduction

In this article we develop a method for clustering evidence in very large scale problems within Dempster–Shafer theory [14, 15, 40–44, 46]. We consider the case when evidence come from multiple events which should be handled independently, and it is not known to which event a piece of evidence is related. We use the clustering process to separate the evidence into subsets for each event, so that each subset may be handled separately.

In an earlier article [38] one of us developed a method using a neural network structure similar to the Hopfield and Tank model [22] for partitioning evidence into clusters for relative large scale problems. All the weights were set a priori using the conflict in Dempster's rule, thus no learning process was utilized. This clustering approach represented a great improvement in computational complexity compared to a previous method based on iterative optimization [29–35], although its clustering performance was not equally good. In order to improve clustering performance a hybrid of the two methods was also developed [36].

In a recent paper [37] this method was further extended for simultaneous clustering and determination of number of clusters during iteration in the neural structure. Here, we let the neuron output signals represent the degree to which pieces of evidence belong to corresponding clusters. From these signals we derive a probability distribution regarding the number of clusters, which gradually during the iteration is transformed into a determination of the number of clusters. This probability distribution is fed back into the neural structure at each iteration to influence the clustering process.

For very large problems we need a new method with still lower computational complexity than achieved so far. This should be attained without sacrificing any of the clustering performance in comparison to the previous neural method [38].

In this article, we combine Dempster–Shafer theory with the antiferromagnetic Potts model [45] into a powerful solver for very large Dempster–Shafer clustering problems. A second aim of the article is purely theoretical; we show how the antiferromagnetic Potts model is isomorphic to a graph optimization problem. This may seem misplaced at first glance, but the variables in the Dempster–Shafer model have a very natural interpretation in terms of the graph model.

The clustering methodology developed over several papers was initially intended for preprocessing of intelligence information for situation analysis in antisubmarine warfare [7, 8]. Having now developed a method with a much lower computational complexity, we believe it can serve as a general solution for preprocessing of intelligence data in information fusion [39].

In Sect. 2 we describe the basics of Dempster–Shafer theory. Section 3 presents the Potts spin model, and the corresponding clustering model is discussed in Sect. 4. Section 5 shows how a linearized Dempster–Shafer conflict function maps to an antiferromagnetic Potts model. The isomorphism between the Potts model and a graph model is developed in Sect. 6, and the resulting relation between the Dempster–Shafer model and the graph model is investigated in Sect. 7. Finally, in Sect. 8 we compare the

M. Bengtsson, J. Schubert (✉)
Department of Data and Information Fusion,
Division of Command and Control Warfare Technology,
Swedish Defence Research Agency, SE-172 90 Stockholm, Sweden
E-mail: schubert@foi.se
url: http://www.foi.se/fusion/

*Present address*:
M. Bengtsson,
P. O. Box 1165, SE-581 11 Linköping, Sweden
E-mail: matben@foi.se

clustering performance and computational complexity of three different clustering methods: the Potts spin clustering developed in this article, a previous developed neural clustering method [38] inspired by Hopfield and Tank [22], and the iterative optimization [29] initially developed.

## 2
## Dempster–Shafer theory

In Dempster–Shafer theory [14, 40] belief is assigned to a proposition by a basic probability assignment. The proposition is represented by a subset $A$ of an exhaustive set of mutually exclusive possibilities, a frame of discernment $\Theta$.

The basic probability assignment (or mass function) is a function from the power set of $\Theta$ to $[0, 1]$

$$m : 2^\Theta \to [0, 1] \tag{1}$$

whenever

$$m(\varnothing) = 0 \tag{2}$$

and

$$\sum_{A \subseteq \Theta} m(A) = 1 \tag{3}$$

where $m(A)$ is called a basic probability number, that is the belief committed exactly to $A$.

The total belief of a proposition $A$ is obtained from the sum of belief for those propositions that are subsets of the proposition in question and the belief committed exactly to $A$

$$\text{Bel}(A) = \sum_{B \subseteq A} m(B) \tag{4}$$

where $\text{Bel}(A)$ is the total belief in $A$ and $\text{Bel}(\cdot)$ is called a belief function

$$\text{Bel} : 2^\Theta \to [0, 1] \ . \tag{5}$$

A subset $A$ of $\Theta$ is called a focal element of Bel if the basic probability number for $A$ is non-zero.

In addition to the belief in a proposition $A$ it is also of interest to know how plausible a proposition might be, i.e., the degree to which we do not doubt $A$. The plausibility,

$$\text{Pls} : 2^\Theta \to [0, 1] \tag{6}$$

is defined as

$$\text{Pls}(A) = 1 - \text{Bel}(A^c) \ . \tag{7}$$

We can calculate the plausibility directly from the basic probability assignment

$$\text{Pls}(A) = \sum_{B \cap A \neq \varnothing} m(B) \ . \tag{8}$$

Thus, while belief in $A$ measures the total mass certainly committed to $A$, plausibility measures the total probability that is in or can be moved into $A$, i.e., $\text{Bel}(A) \leq \text{Pls}(A)$.

If we receive a second item of information concerning the same issue but from a different source, the two items can be combined to yield a more informed view. Combining two belief functions is done by calculating the orthogonal combination using Dempster's rule. This is most simply illustrated by the combination of basic probability assignments. Let $A_i$ be a focal element of $\text{Bel}_1$ and let $B_j$ be a focal element of $\text{Bel}_2$. Combining the corresponding basic probability assignments $m_1$ and $m_2$ results in a new basic probability assignment $m_1 \oplus m_2$

$$m_1 \oplus m_2(A) = K \cdot \sum_{A_i \cap B_j = A} m_1(A_i) m_2(B_j) \tag{9}$$

where $K$ is a normalizing constant, $K = 1/(1 - \kappa)$ where

$$\kappa = \sum_{A_i \cap B_j = \varnothing} m_1(A_i) m_2(B_j) \ . \tag{10}$$

This normalization is needed since, by definition, no mass may be committed to $\varnothing$ (according to standard theory [40]; for an alternative view, an "open world assumption", see [44]). The new belief function $\text{Bel}_1 \oplus \text{Bel}_2(\cdot)$ can be calculated by the above formula from $m_1 \oplus m_2(\cdot)$. When we wish to combine several belief functions this is done simply by combining the first two, then combining the result with the third and so forth.

The weight of conflict between two belief functions is defined as $\text{Con}(\text{Bel}_1, \text{Bel}_2) = -\log(1 - \kappa)$. This function will be the starting point in our analysis of evidence clustering.

## 3
## Potts spin

The physics of spin systems is a quite different area in science which will be utilized here. The Hopfield model [21], based on the physics of Ising spins (see for instance [16] for a review), was the first model to bridge the gap between spin systems and computer science, and gained a wider interest. The Potts model (see for instance [45]) is a generalization of the Ising model where each spin may have an arbitrary (but finite) degree of freedom instead of just two. It has proven useful in many complex optimization problems [19, 20, 24, 27].

If the Potts spin at site $i$ is denoted $\sigma_i = 1, 2, \ldots, q$, where $q$ is a positive integer, the energy function that defines the model is written in terms of spin–spin interactions,

$$E = \frac{1}{2} \sum_{i,j=1}^N J_{ij} \delta_{\sigma_i, \sigma_j} \ . \tag{11}$$

Another useful notation is to treat each Potts spin as a discrete vector in a hypercube: $S_{ia} = 0, 1$ with the constraint $\sum_{a=1}^q S_{ia} = 1 \ \forall \ i$ where $a$ is the vector index. Then the energy function becomes

$$E = \frac{1}{2} \sum_{i,j=1}^N \sum_{a=1}^q J_{ij} S_{ia} S_{ja} = \frac{1}{2} \sum_{i,j=1}^N J_{ij} \vec{S}_i \circ \vec{S}_j \ , \tag{12}$$

where the vector character of the spin has been used in the last equation. The spins merely encode which class a data (point) belongs to; $S_{ia} = 1$ means that the site $i$ belongs to class $a$.

Here we assume that the interaction is completely antiferromagnetic ($J_{ij} \geq 0$). This model can serve as a data clustering algorithm with a spin on each data point (site), if $J_{ij}$ is used as a penalty factor of site $i$ and $j$ being in the same class; sites in different classes get no penalty. So far, the spins are only a formalism to describe the problem in other terms, and the problem consists of minimizing an energy function by flipping the spins into different states. This spin flipping process takes place via simulated annealing, whereby the complete spin system is viewed as being contained in a thermal reservoir at a certain temperature. At a high temperature, the spins flip more or less at random, and are only marginally biased by their interactions ($J_{ij}$). As the temperature is lowered, discontinuous phenomena such as phase transitions may occur. During a phase transition, parts of the system become constrained in one way or the other, they freeze. Finally, when the complete system is frozen, the spins are completely biased by the interactions ($J_{ij}$) so that, hopefully, the minimum of the energy function is reached.

In a Monte Carlo simulation, the spins are actual stochastic states, and such a simulation is usually very time consuming. Instead, we will use a mean field model, where spins are deterministic [27], which is usually very time efficient. The Potts model has been used for various forms of clustering data [1, 2, 4–6, 9–13, 28], and other complex optimization problems [23, 25, 26].

## 4
## The problem

If we receive several pieces of evidence about different and separate events, and the pieces of evidence are mixed up, our task is to arrange them according to which event they are referring to. Thus, we partition the set of all pieces of evidence $\chi$ into subsets where each subset refers to a particular event. In Fig. 1 these subsets are denoted by $\chi_i$. The conflict when all pieces of evidence in $\chi_i$ are combined by Dempster's rule is denoted by $c_i$. Here, thirteen pieces of evidence are partitioned into four subsets. When the number of subsets is uncertain there will also be a "domain conflict" $c_0$ which is a conflict between the current hypothesis about the number of subsets and our prior belief, although this situation is not considered in this article. The partition is then simply an allocation of all pieces of evidence to the different events. Since pieces of evidence corresponding to different events are unrelated, the evidence belonging to a particular event can be analyzed within the Dempster–Shafer paradigm, independently of any other pieces of evidence.

If it is not known a priori which event each piece of information is corresponding to, we have a problem. It could then be impossible to know directly if two different pieces of evidence are corresponding to the same event. We do not know if we should put them into the same subset or not. It is precisely this problem we are facing in this article. The task is to organize the evidence into different classes so that each class precisely corresponds to a particular event in a unique way.

To solve this problem, we can use the conflict in Dempster's rule when all pieces of evidence within a subset are combined, as an indication of whether these pieces of evidence belong together. The higher this conflict is, the less credible that they belong together.

Let us create an additional piece of evidence for each subset with the proposition that this is not an "adequate partition". We have a simple frame of discernment on the metalevel $\Theta = \{AdP, \neg AdP\}$, where AdP is short for "adequate partition." Let the proposition take a value equal to the conflict of the combination within the subset,

$$m_{\chi_i}(\neg AdP) \triangleq \text{Conf}(\{S_j | S_j \in \chi_i\}) \ . \tag{13}$$

These new pieces of evidence (simple support functions), one regarding each subset, reason about the partition of the original evidence. Just so that they are not confused with the original evidence $\{S_j\}$, let us call this evidence "metalevel evidence," and let us say that its combination and the analysis of that combination take place on the "metalevel," Fig. 1.

In [29] a criterion function of overall conflict called the metaconflict function for reasoning with multiple events was established. The metaconflict is derived as the plausibility that the partitioning is correct when the conflict in each subset is viewed as a piece of metalevel evidence against the partitioning of the set of evidence, $\chi$, into the subsets, $\chi_i$.

**Definition** *Let the* **metaconflict function,**

$$Mcf(q, S_1, S_2, \ldots, S_n) \triangleq 1 - (1 - c_0) \prod_{i=1}^{q} (1 - c_i) \ , \tag{14}$$

*be the conflict against a partitioning of n pieces of evidence of the set $\chi$ into q disjoint subsets $\chi_i$. Here, $c_i$ is the conflict in subset i and $c_0$ is the conflict between q subsets and propositions about possible different numbers of subsets.*

We will use the minimizing of the metaconflict function as the method of partitioning the evidence into subsets corresponding to the events. This method will also handle the situation when the number of events are uncertain. After this, each subset refers to a different event and the reasoning can take place with each event treated separately.

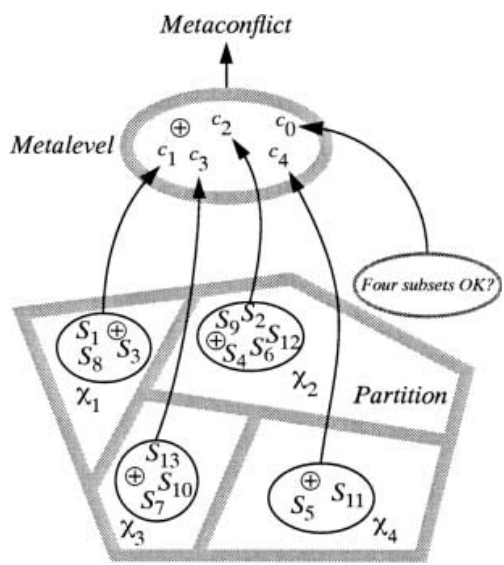

**Fig. 1.** The conflict in each subset of the partition becomes a piece of evidence at the metalevel

## 5
## From Dempster–Shafer Theory to Potts Spin

The metaconflict function is easier to treat if it is rewritten as a sum instead of a product, by taking the logarithm. The number of clusters is here kept unchanged ($c_0$ is constant). Let us rewrite the minimization as follows

$$\min Mcf = \min\left[1 - \prod_i (1 - c_i)\right]$$

$$\Leftrightarrow$$

$$\max(1 - Mcf) = \max \prod_i (1 - c_i)$$

$$\Leftrightarrow$$

$$\max \log(1 - Mcf) = \max \log \prod_i (1 - c_i)$$

$$= \max \sum_i \log(1 - c_i) = \min \sum_i -\log(1 - c_i) \tag{15}$$

where $-\log(1 - c_i) \in [0, \infty]$ is a weight [40, p. 77] of evidence, i.e., in this context a weight of conflict.

Since the minimum of Mcf (=0) is obtained when the final sum is minimal (=0), the minimization of the final sum yields the same result as a minimization of Mcf would have done.

In Dempster–Shafer theory one defines a simple support function, where the evidence points precisely and unambiguously to a single nonempty subset $A$ of $\Theta$. If $S$ is a simple support function focused on $A$, then the basic probability numbers are denoted $m(A) = s$, and $m(\Theta) = 1 - s$. If two simple support functions, $S_1$ and $S_2$ [40, p. 95], focused on $A_1$ and $A_2$ respectively, are combined, the weight of conflict between them is [40]

$$\text{Con}(S_1, S_2) = \begin{cases} -\log(1 - s_1 s_2) & \text{if } A_1 \cap A_2 = \varnothing \\ 0, & \text{else} \end{cases}, \tag{16}$$

which may be written as

$$\text{Con}(S_1, S_2) = -\log(1 - s_1 s_2)\delta_{|A_1 \cap A_2|} \tag{17}$$

with $\delta_{|A_1 \cap A_2|}$ being defined so that it is unity for $A_1 \cap A_2 = \varnothing$ and zero otherwise.

Generally, the conflict function of a $(N + 1)$-system can be obtained recursively from a $(N)$-system via [40]

$$\text{Con}(S_1, S_2, \ldots, S_{N+1})$$
$$= \text{Con}(S_1, S_2, \ldots, S_N)$$
$$+ \text{Con}(S_1 \oplus S_2 \oplus \cdots \oplus S_N, S_{N+1}), \tag{18}$$

where $S_1 \oplus S_2$ etc. denotes the combination of two simple support functions via Dempster's rule of combination.

Conventional algorithms, and the Potts model in particular, handle data as pairwise terms. It is therefore necessary to simplify the conflict function, and write the conflict of a sum of support functions as the sum of pairwise conflicts, i.e., to linearize the conflict function. The approximation,

$$\text{Con}(S_1 \oplus S_2 \oplus \cdots \oplus S_N, S_{N+1})$$
$$\approx \text{Con}(S_1, S_{N+1})$$
$$+ \text{Con}(S_2, S_{N+1}) + \cdots + \text{Con}(S_N, S_{N+1}), \tag{19}$$

is precisely such a linearization.

Minimizing a sum of $-\log(1 - s_j s_k)$ terms is as an approximation correct to leading order, i.e., all second order terms in $\{s_i\}$ are unchanged in this approximation. The actual function being minimized is

$$\sum_i \sum_{\substack{k,l \\ S_k, S_l \in \chi_i}} -\log(1 - s_k s_l)$$

$$= \sum_i -\log \prod_{\substack{k,l \\ S_k, S_l \in \chi_i}} (1 - s_k s_l)$$

$$= \sum_i -\log\left[1 - \left(\sum_{\substack{k,l \\ S_k, S_l \in \chi_i}} s_k s_l - Y\right)\right] \tag{20}$$

while the function above, Eq. (15), can be rewritten as

$$\sum_i -\log(1 - c_i) = \sum_i -\log\left[1 - \left(\sum_{\substack{k,l \\ S_k, S_l \in \chi_i}} s_k s_l - X\right)\right] \tag{21}$$

where $X$ and $Y$ are the higher order terms.

These functions are identical in their first order terms, and $Y \leq X$. Hence, all leading order terms are correct, and nonleading order terms are suppressed by powers of $s$, which are all smaller than one. Moreover, the conflict of a purely conflict free configuration is zero in both cases. Thus, the actual minimization slightly overestimates the conflict within the subset.

The linearized conflict function is therefore written recursively as

$$\text{ConL}(S_1, S_2, \ldots, S_{N+1}) = \text{ConL}(S_1, S_2, \ldots, S_N)$$
$$- \sum_{i=1}^{N} \log(1 - s_i s_{N+1})\delta_{|A_i \cap A_{N+1}|}, \tag{22}$$

which is in an appropriate form to be identified with the Potts model.

Minimizing the weight of conflict within each cluster is equivalent to maximizing the weight of conflict between each cluster, as will be shown below (Eq. (26)).

We denote the cluster an evidence belongs to with $\chi_i$, which may be any positive integer between one and an upper limit $K$. We will always use $q = K$. If two pieces of evidence have disjoint cores ($A_1 \cap A_2 = \varnothing$), they need not be in conflict if they refer to two different events; only when they refer to the same event are they in conflict. Hence, the recursive relation for the conflict function is modified to:

$$\text{ConL}(S_1, S_2, \ldots, S_{N+1})$$
$$= \text{ConL}(S_1, S_2, \ldots, S_N)$$
$$- \sum_{i=1}^{N} \log(1 - s_i s_{N+1}) \delta_{|A_i \cap A_{N+1}|} \delta_{\chi_i \chi_{N+1}} \quad , \qquad (23)$$

where it remains to determine $\{\chi_i\}_{i=1}^{k}$. Note that the interaction $(-\log(1 - s_i s_{N+1}) \delta_{|A_i \cap A_{N+1}|})$ is fixed for any given problem. The Potts model, as defined by Eq. (11), maps naturally to clustering problems. If $J_{ij} \geq 0, \forall\, i, j$, the model is said to be antiferromagnetic, i.e., any pair of spins in the same state are repellent. It has therefore qualitatively the same type of interaction as the conflict function; both are repellent.

In terms of clustering, it describes so called parametric clustering, where data is centered around a fictitious class prototype. The isomorphism between Eq. (23) and the Potts model is seen if the Potts energy function is expressed recursively as

$$E_{N+1} = E_N + \sum_{i=1}^{N} J_{i,N+1} \delta_{\sigma_i \sigma_{N+1}} \qquad (24)$$

and the spin state, $\sigma_i, 1 \leq \sigma_i \leq q$, is identified with the class index $\chi_i$. Thus, each piece of evidence is translated into a Potts spin. Moreover, the spin–spin interaction $J_{i,N+1} = -\log(1 - s_i s_{N+1}) \delta_{|A_i \cap A_{N+1}|}$, is valid for any pair of indices $i$ and $j$. The relations between the Potts model and the linearized conflict function are summarized as:

$$\begin{aligned} E_N &\leftrightarrow \text{ConL}(S_1, S_2, \ldots, S_N) \\ J_{ij} &\leftrightarrow -\log(1 - s_i s_j) \delta_{|A_i \cap A_j|} \\ \sigma_i &\leftrightarrow \chi_i \\ q &\leftrightarrow K \end{aligned} \qquad (25)$$

If we instead seek to maximize the weight of conflict between each class, then $\sum_{ij} J_{ij}(1 - \delta_{\sigma_i \sigma_j})$ should be maximized, or alternatively, minimize

$$\begin{aligned} E &= -\frac{1}{2} \sum_{ij} J_{ij}(1 - \delta_{\sigma_i \sigma_j}) \\ &= -\frac{1}{2} \sum_{ij} J_{ij} + \frac{1}{2} \sum_{ij} J_{ij} \delta_{\sigma_i \sigma_j} \quad . \end{aligned} \qquad (26)$$

The term $\sum_{ij} J_{ij}$ is just a constant, and consequently, maximization of the weight of conflict between each class is equivalent to the minimization of the weight of conflict within each class.

However, the isomorphism between the linearized Dempster–Shafer clustering model and the antiferromagnetic Potts spin model does not solve the problem by itself, but by virtue of the dynamics of the spins. As discussed earlier, statistical mechanics of the Potts spin in a mean field approximation will supply the necessary dynamics. Using simulated annealing, the temperature acts as a control parameter that, via a sequence of phase transitions, resolves finer and finer details of the data; an algorithmic structure that is well suited for clustering problems.

Figure 2 illustrates parametric clustering of 2D data. While nonparametric clustering represents the most constraint free form of clustering, and consequently, benchmark studies are difficult to perform, parametric clustering has a well defined energy function to be optimized. For sparsely distributed Gaussian clusters of the same size, analytical estimates of the optimal Bayes limit can be made, which agree with fairly large precision with numerical results of the mean field Potts model [4]. Numerical results for Dempster–Shafer clustering are discussed in Sect. 8.

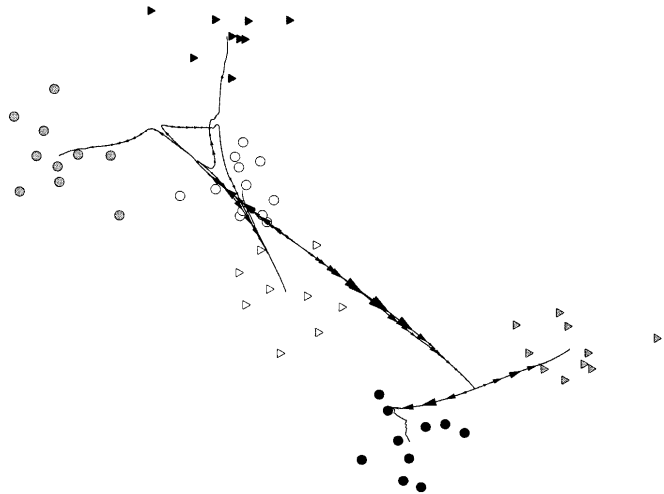

**Fig. 2.** Six similar Gaussian distributions in 2D are clustered using the Potts spin model. Arrows indicate cluster centers ($\mathbf{y}_a = (\sum_i x_i V_{ia})/(\sum_i V_{ia})$, see Sect. 7 for the notation)

The fact that there are more than one global minimum is, in contrast to naïve expectations, not a feature that makes the problem simpler to solve, quite the opposite. During simulated annealing, when the temperature is lowered and a phase transition occurs, one corner of the system may freeze partly into a state corresponding to one of the global minima, while another corner of the system freezes into another global minimum state. The border between these two (or more) states, is most certainly going to give rise to additional energy, which results in a non-optimal solution. This is called frustration, and has been studied extensively in highly frustrated spin glass models of Ising spins (see [16] for a review). It is well known that mean field theory is not suitable to study spin glass models, but as will be seen below, acceptable solutions are found in most cases for the Dempster–Shafer clustering problems. This indicates that despite the degenerate ground state, frustration is not a too difficult problem here.

Using the vector notation for the Potts spin, the complete energy function we are considering is

$$\begin{aligned} E[S] = &\frac{1}{2} \sum_{a=1}^{K} \sum_{i,j=1}^{N} J_{ij} S_{ia} S_{ja} - \frac{\gamma}{2} \sum_{a=1}^{K} \sum_{i=1}^{N} S_{ia}^2 \\ &+ \frac{\alpha}{2} \sum_{a=1}^{K} \left( \sum_{i=1}^{N} S_{ia} \right)^2 \end{aligned} \qquad (27)$$

where the first term is the standard clustering cost. The $\gamma$-term, which does not influence the relative energy levels, is necessary to avoid a bifurcating behavior. It simply adds a constant to the interactions that moves any potentially harmful negative eigenvalues along the positive axis. In all simulations, $\gamma = 0.5$. The $\alpha$-term is an energy cost with an equipartition effect; it has its minimum for configurations with an equal number of items in each cluster, and increases when there is an unbalance between the clusters. This is done to favor those balanced solutions that are intuitively appealing among all different global optima.

The Potts mean field equations are [27]:

$$V_{ia} = \frac{e^{-H_{ia}[V]/T}}{\sum\limits_{b=1}^{K} e^{-H_{ib}[V]/T}} \qquad (28)$$

where

$$H_{ia}[V] = \frac{\partial E[V]}{\partial V_{ia}} = \sum_{j=1}^{N} J_{ij}V_{ja} - \gamma V_{ia} + \alpha \sum_{j=1}^{N} V_{ja} \qquad (29)$$

with $V_{ia} = \langle S_{ia} \rangle$ (thermal averages), and are used recursively (serially) until a stationary equilibrium state has been reached for each temperature. To apply it to Dempster–Shafer clustering we use interactions $J_{ij} = -\log(1 - s_i s_j)\delta_{|A_i \cap A_j|}$.

The algorithm for simulating these spins works roughly as follows. Use a precomputed highest critical temperature, $T_c$, as the starting temperature. Choose the mean field spins to be in their symmetric high temperature state; $V_{ia} = 1/K \,\forall\, i, a$. At each temperature, iterate Eqs. (28, 29) until a fix point has been reached. The temperature is lowered by a constant factor until every spin has frozen, i.e., $V_{ia} = 0, 1$, Fig. 3.

Problematic events not seldom have one completely empty cluster, but by using the equipartition term (the $\alpha$-term) more balanced partitions are obtained at the risk of introducing spurious minima. Especially for large problem sizes it proves useful to have a small $\alpha$-term.

# 6
# Potts and graph isomorphism

There exists a known isomorphism between a ferromagnetic Potts model and a graph model [18]. While the spins are placed on the sites (vertices, nodes), and have $q$ degrees of freedom, the graph model has bond variables placed on the edges (links) between the sites. A bond variable between site $i$ and $j$, $n_{ij}$, may be on (occupied), $n_{ij} = 1$, or off (vacant), $n_{ij} = 0$. To each edge is associated an a priori given probability, $0 \le p_{ij} \le 1$, which roughly tells how likely it is for a bond being on. Both the ferromagnetic Potts model and the corresponding graph model may in fact be obtained as the marginal distributions of a combined model, containing spins as well as dynamic bond variables [17]. Hence, a dual description of the model is possible, either in terms of spins, or in terms of edges, or both.

A ferromagnetic spin system is poorly suited for deterministic clustering since the ground state is completely degenerated with all spins being parallel. Adding a weak, non-site-dependent, antiferromagnetic term to the ferromagnetic system, this degeneracy is broken, and a nonparametric clustering solution may be obtained from deterministic annealing [5]. There exists an isomorphic graph model also to this Potts model [3], but it is slightly different from the previous one.

However, the Potts model relevant in this study is completely *antiferromagnetic* ($J_{ij} \ge 0 \,\forall\, i, j$), and the task is to find the corresponding graph model. The probability distribution of the spin configurations is given by the Boltzmann factor,

$$\mathscr{P}_P(\{\sigma\}) = \frac{1}{Z}e^{-\beta E} = \frac{1}{Z}\exp\left(-\beta \sum_{(ij)} J_{ij}\delta_{\sigma_i \sigma_j}\right)$$

$$= \frac{1}{Z}\prod_{(ij)}\exp(-\beta J_{ij}) \qquad (30)$$

with *(ij)* meaning every unique pair of sites, and where $Z$ is the usual partition function that acts as a normalization:

---

**INITIALIZE**
$K$ (the problem size); $N = 2^K - 1$;
$J_{ij} = -\log(1 - s_i s_j)\,\delta_{|Ai \cap Aj|}\;\; \forall i, j$;
$s = 0;\; t = 0;\; \varepsilon = 0.001;\; \tau = 0.9;\; \alpha$ (for $K \le 7$: $\alpha = 0$, $K = 8$: $\alpha = 10^{-6}$, $K = 9$: $\alpha = 0$, $K = 10$: $\alpha = 3 \cdot 10^{-7}$, $K = 11$: $\alpha = 3 \cdot 10^{-8}$); $\gamma = 0.5$;
$T^0 = T_c$ (a critical temperature) $= \frac{1}{K} \cdot max(-\lambda_{min}, \lambda_{max})$, where $\lambda_{min}$ and $\lambda_{max}$ are the extreme eigenvalues of M, where $M_{ij} = J_{ij} + \alpha - \gamma\,\delta_{ij}$;
$V_{ia}^0 = \frac{1}{K} + \varepsilon \cdot rand[0,1]\;\; \forall i, a$;

**REPEAT**
- **REPEAT–2**
  $\forall i$ Do:
  - $H_{ia}^s = \sum\limits_{j=1}^{N}(J_{ij} + \alpha)V_{ja}^s - \gamma V_{ia}^s\;\; \forall a$;

  - $F_i^s = \sum\limits_{a=1}^{K} e^{-H_{ia}^s/T^t}$;

  - $V_{ia}^{s+1} = \frac{e^{-H_{ia}^t/T^t}}{F_i^s} + \varepsilon \cdot rand[0,1]\;\; \forall a$;

  - $s = s + 1$;

  **UNTIL–2**
  $\frac{1}{N}\sum\limits_{i,a}\left|V_{ia}^s - V_{ia}^{s-1}\right| \le 0.01$;

- $T^{t+1} = \tau \cdot T^t$;
- $t = t + 1$;

**UNTIL**
$\frac{1}{N}\sum\limits_{i,a}(V_{ia}^s)^2 \ge 0.99$;

**RETURN**
$\{\chi_a | \forall S_i \in \chi_a \cdot \forall b \ne a\; V_{ia}^s > V_{ib}^s\}$;

**Fig. 3.** The clustering algorithm. The saturation $\frac{1}{N}\sum_{i,a}(V_{ia}^t)^2$ measures how "frozen" the system is (the specific numerical values given are just examples)

$$Z = \sum_{\{\sigma\}} e^{-\beta E(\{\sigma\})} \qquad (31)$$

The distribution, $\mathscr{P}_P(\{\sigma\})$, is rewritten as

$$\mathscr{P}_P(\{\sigma\}) = \frac{1}{Z} \prod_{(ij)} \left[ e^{-\beta J_{ij}} + \left(1 - e^{-\beta J_{ij}}\right)\left(1 - \delta_{\sigma_i \sigma_j}\right) \right] . \qquad (32)$$

Note that the product may be interpreted as a product over each edge in a graph. At each edge, define a probability $p_{ij} = 1 - e^{-\beta J_{ij}}$, which we denote the bond probability. In this clustering model, all interactions are repellent and increase with "distance." Thus, data that are in some sense "close," with small $J_{ij}$, have also a small bond probability, and vice versa. The distribution with $p_{ij}$'s is

$$\mathscr{P}_P(\{\sigma\}) = \frac{1}{Z} \prod_{(ij)} \left[ (1 - p_{ij}) + p_{ij}\left(1 - \delta_{\sigma_i \sigma_j}\right) \right] . \qquad (33)$$

The combined spin and bond model is defined as

$$\mathscr{P}_{PG}(\{n\}, \{\sigma\})$$
$$= \frac{1}{Z} \prod_{(ij)} \left[ (1 - p_{ij})\delta_{n_{ij}1} + p_{ij}\delta_{n_{ij}0}\left(1 - \delta_{\sigma_i \sigma_j}\right) \right] , \qquad (34)$$

with dynamic spins ($\sigma_i = 1, 2, \ldots, q$) on the sites, as well as dynamic bond variables ($n_{ij} = 0, 1$) on the links *between* the sites. It is simple to verify that the marginal distribution of $\mathscr{P}_{PG}\{n\}, \{\sigma\}$ obtained by summing over all possible bond states, is indeed identical to the Potts spin distribution $\mathscr{P}_P(\{\sigma\})$,

$$\sum_{\{n\}} \mathscr{P}_{PG}(\{n\}, \{\sigma\}) = \mathscr{P}_P(\{\sigma\}) . \qquad (35)$$

It is also a simple matter to find the conditional distribution $\mathscr{P}_{PG}(\{n\}|\{\sigma\})$ from Eq. (34); just read off $\mathscr{P}_{PG}(\{n\}, \{\sigma\})$ to get:

$$\sigma_i = \sigma_j \Rightarrow n_{ij} = 1$$
$$\sigma_i \neq \sigma_j \Rightarrow \begin{cases} n_{ij} = 0, & \text{with probability } p_{ij} \\ n_{ij} = 1, & \text{with probability } 1 - p_{ij} \end{cases} . \qquad (36)$$

Hence, parallel spins always have a bond being *on* on the edge connecting the spins, while nonparallel spins have the corresponding bond *off* with the bond probability.

Finding the corresponding graph model, obtained by summing over all spin configurations of $\mathscr{P}_{PG}(\{n\}, \{\sigma\})$, is more difficult since it does not factor in terms of the spins. Collect factors containing bonds being on, and bonds being off, respectively,

$$\mathscr{P}_{PG}(\{n\}, \{\sigma\})$$
$$= \frac{1}{Z} \left[ \prod_{(ij):n_{ij}=1} (1 - p_{ij}) \right] \left[ \prod_{(ij):n_{ij}=0} p_{ij}\left(1 - \delta_{\sigma_i \sigma_j}\right) \right] . \qquad (37)$$

At this stage introduce the notion of a *spin cluster*, which is defined as a maximal set of sites with identical (parallel) spins. On the other hand a *bond cluster* is a maximal nonempty subset of occupied bonds, where there is a path of occupied bonds between any two bonds in the bond cluster (a bond-cluster of a graph is a maximal nonempty connected subgraph [18]). To get a relatively simple expression for the distribution of the bond model it is necessary to assume *full connectivity*, i.e., between any two sites there exists a single edge that directly connects them. In the words of [18], full connectivity means that between any two vertices (sites) there is a path with only a single edge.

The interesting quantity in this computation is

$$\sum_{\{\sigma_i\}_{i=1}^N} \prod_{(ij):n_{ij}=0} p_{ij}\left(1 - \delta_{\sigma_i \sigma_j}\right) , \qquad (38)$$

which contains essentially two pieces of information. First, any vacant bond must have nonidentical spins at its two sites. This follows also directly from Eq. (36). Since two different bond-clusters are connected only by vacant bonds, then any spin in one bond-cluster must be different from any spin in another bond-cluster. Secondly, the sum over spins gives a combinatorial weight of the states that is generally different for different configurations. For instance, if the constraint $(1 - \delta_{\sigma_i \sigma_j})$ is removed, then the sum above simply reduces to

$$q^N \prod_{(ij):n_{ij}=0} p_{ij} \qquad (39)$$

Hence, the actual combinatorial factor for any state is smaller than $q^N$.

Since the number of spin degrees of freedom is limited to $q$, and a spin state in one bond-cluster is never allowed in another bond-cluster, the number of bond-clusters, $C(\{n\})$, is thus limited to $q$:

$$1 \leq C(\{n\}) \leq q . \qquad (40)$$

Note that this result relies on the full connectivity condition, and that an upper limit on $C(\{n\})$ may not exist without this condition.

In Eq. (38), the sum over spins contains every site, while the product contains only vacant bonds. It is difficult to find a general result, so we will start with a simplification. Assume for the moment that there are no vacant bonds between sites in the same bond-cluster, so that all vacant bonds connect sites from different bond-clusters. The $q$ spin directions must be partitioned into $C(\{n\})$ nonempty sets. Denote the number of available spin directions in bond-cluster $a$ with $q_a$ (constrained to $q = \sum_{a=1}^C q_a$), and the number of sites $N_a$. For each partition of the spin directions among the bond-clusters, sum over all possible spin configurations. The number of such configurations of a bond-cluster $a$ is $q_a^{N_a}$, since there are no other constraints on the spins within a bond-cluster. Moreover, there are $q!/(q_1!q_2! \cdots q_c!)$ ways to partition the available spin directions among the $C$ bond-clusters. With our previous assumptions, the sum over spins is now written,

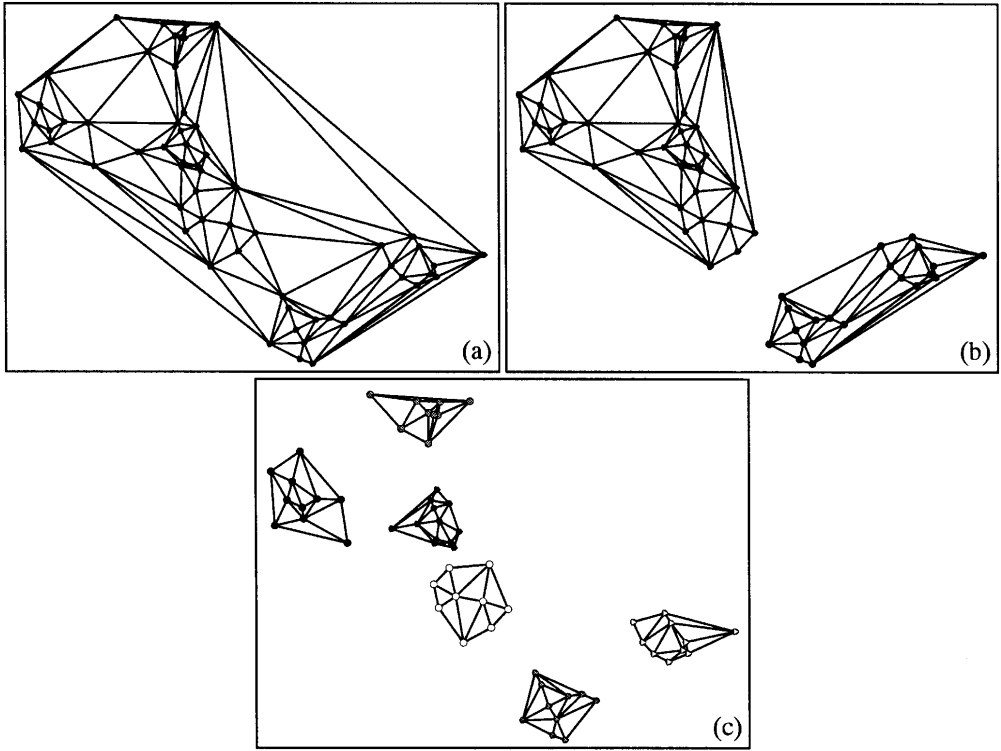

**Fig. 4.** This is an illustration of how bond-clusters are formed during annealing. For simplicity, only a triangulation of the bonds in each bond-cluster is shown. Bond-clusters are chosen deterministically using the condition $p_{ij} > 0.5$

$$\sum_{\{\sigma_i\}_{i=1}^{N}} \prod_{(ij):n_{ij}=0} p_{ij}\left(1 - \delta_{\sigma_i\sigma_j}\right)$$

$$\approx \left( \sum_{\substack{q_1,q_2,\cdots,q_c \\ q=\Sigma_a q_a}}^{q} \frac{q!}{q_1!q_2!\cdots q_c!} q_1^{N_1} q_2^{N_2} \cdots q_c^{N_c} \right)$$

$$\times \left( \prod_{(ij):n_{ij}=0} p_{ij} \right) . \tag{41}$$

It may seem to be an error that the upper summation limit is $q$ and not, for instance, $q - C$. However, illegal configurations with any $q_a = 0$ contribute zero due to a factor $q_a^{N_a} \rightarrow 0^{N_a} = 0$.

The fact that vacant bonds may connect sites in the same bond-cluster is not taken into account in the previous analysis. For instance, if bond-cluster no. 1 contains a single vacant bond, the number of possible spin states is reduced from $q_1^{N_1}$ to $(q_1 - 1)q_1^{(N_1-1)}$. Or if there are two vacant bonds, that do not have any sites in common in a cluster, the combinatorial factor for that cluster becomes $(q_1 - 1)^2 q_1^{(N_1-2)}$. Both these cases have an additional constraint $q_1 \geq 2$. Although these combinatorial factors are all less than $q_1^{N_1}$, the number of possible states with, for instance, a single vacant bond inside a bond-cluster, is in general large. For this reason alone, these configurations cannot be neglected.

It turns out that the configurations with vacant bonds inside bond-clusters disappear when the ground state is reached for the following reason. The ground state of the spin system consists of $q$ localized spin-clusters,[1] where all spins are parallel within each cluster. Because $\sigma_i = \sigma_j \Rightarrow n_{ij} = 1$ according to Eq. (36), all bonds connecting sites in the same spin-cluster are occupied in the ground state of the system.

---

[1] This can be seen directly from Eq. (11), which has a minimum value that is a decreasing function with respect to $q$.

The question is then if there are any occupied bonds connecting different spin-clusters. Since the $J_{ij}$'s are the independent interactions, and the $p_{ij}$'s the dependent ones, letting $T \rightarrow 0$ gives $p_{ij} \rightarrow 1$. A bond connecting different spin-clusters ($\sigma_i \neq \sigma_j$) is vacant with probability $p_{ij}$ according to Eq. (36). Hence, for $T \rightarrow 0$, the probability for vacant bonds between spin-clusters goes to unity, and are thus highly unlikely to be occupied. Therefore, the ground state of a spin/bond model is characterized by $q$ localized spin-clusters, where the bond-clusters are identical (contain the same set of vertices) to the spin-clusters, and no bonds are vacant inside bond-clusters.

So, Eq. (41) is expected to be correct for $T \rightarrow 0$, which gives the total bond state distribution

$$\mathscr{P}_G(\{n\}; C(\{n\}))$$

$$= \frac{1}{Z} \left( \sum_{\substack{q_1,q_2,\cdots,q_c \\ q=\Sigma_a q_a}}^{q} \frac{q!}{q_1!q_2!\cdots q_c!} q_1^{N_1} q_2^{N_2} \cdots q_C^{N_c} \right)$$

$$\times \left( \prod_{(ij):n_{ij}=0} p_{ij} \right) \cdot \left[ \prod_{(ij):n_{ij}=1} (1 - p_{ij}) \right] , \tag{42}$$

with the condition $C(\{n\}) \leq q$.

It may be interesting to make a qualitative analysis of the temperature dependency of the bond model, without any references to the spin model. It is important to understand that, since $\beta$ is a free parameter, $\{J_{ij}\}$ must be treated as independent parameters, and $\{p_{ij}\}$ as dependent. For

$T \to \infty$, $p_{ij} \to 0 \ \forall i,j$ (iff $J_{ij} > 0$). Therefore, $n_{ij} = 1 \ \forall i,j$, and the whole system is one big bond-cluster.

As the temperature is decreased towards zero, $p_{ij} \to 1 \ \forall i,j$, and consequently all bonds ought to be vacant. However, the condition $C(\{n\}) \le q$, which is a remnant from the spin model, implies that not every bond can be vacant. Crudely speaking, we seek max $\#(n_{ij} = 0)$ subject to the constraint $C(\{n\}) \le q$. It is easy to see that this implies $C(\{n\}) = q$. Thus, a single large bond-cluster at a high temperature is broken down to smaller bond-clusters as the temperature is lowered, and eventually $q$ bond-clusters remain as the ground state. In Fig. 4 we have illustrated this behavior, and made a triangulation of sites belonging to identical bond-clusters based on the condition $p_{ij} > 0.5$.

This behavior is quite opposite from the bond-model corresponding to a *ferromagnetic* Potts spin system, where all bonds are vacant at a high temperature, and all are occupied at the ground state. The modified ferromagnetic Potts clustering model, with a weak constant and antiferromagnetic term added [3], need not produce a single large bond-cluster as its ground state, but rather separate bond-clusters representing the underlying data structure.

One issue needs some more attention: the condition of full connectivity. What happens when this condition is relaxed? If $J_{ij} = 0$ for some $i,j$, then there is no direct interaction between $\sigma_i$ and $\sigma_j$, no link from $i$ to $j$, and consequently, there does not exist a bond variable $n_{ij}$. A quantitative analysis depends on the particular graph, and is therefore not practical to perform. Qualitatively it is easier to see what happens. If the graph is sparse enough, there may be sets of sites with no direct interaction between them. For instance, if there exists two non-empty sets of sites without any direct contact, the spins may be parallel without affecting the energy of the system. If the definition of a spin-cluster is adjusted to mean sets of parallel spins that *do* have an interaction between them, this implies that sets of parallel spins may form separate spin-clusters if they are isolated from each other. Thus, $q$ need not set the upper limit of the number of spin-clusters. Moreover, the probability distribution for the bonds (Eq. (42)) will be modified with different combinatorial factors that depend on the underlying graph.

## 7
### Dempster–Shafer clustering and the graph model
We have so far established a connection between the linearized conflict function in Dempster–Shafer theory, and the antiferromagnetic Potts spin model. Also, an isomorphism between this Potts model and a graph optimization problem was described in the previous section. The next step is to find out how the Dempster–Shafer problem and the graph model are related. In principle, we have already demonstrated an isomorphism between the linearized Dempster–Shafer problem and the graph optimization problem. However, if a slight modification to the conflict function is made, a much nicer relation can be derived.

The original way of defining the weight of evidence $w$, corresponding to the degree of support $s$ for an event $A$, is $s = 1 - e^{-w}$ [40, p. 78]. There is an amount of arbitrariness in this definition since it is perfectly allowed to use $s = 1 - e^{-\lambda w}$, where $\lambda$ is an arbitrary positive constant. For no specific reason, $\lambda$ was originally chosen to be one [40]. The same argument applies to the weight of conflict, which may now be written

$$\text{Con}(\text{Bel}_1, \text{Bel}_2) = -\frac{1}{\lambda} \log(1 - \kappa) \tag{43}$$

with $\lambda$ being a positive constant.

The derivation to show the relation between the Dempster–Shafer model, with $\lambda$ included, and the Potts model parallels the previous derivation in Sect. 5. The recursively written and linearized conflict function is now

$$
\begin{aligned}
&\text{ConL}(S_1, S_2, \ldots, S_{N+1}) \\
&= \text{ConL}(S_1, S_2, \ldots, S_N) \\
&\quad - \frac{1}{\lambda} \sum_{i=1}^{N} \log(1 - s_i s_{N+1}) \delta_{|A_i \cap A_{N+1}|} \delta_{\chi_i \chi_{N+1}} \ .
\end{aligned}
\tag{44}
$$

Comparing with the Potts model, Eq. (24), we make the following identifications:

$$
\begin{aligned}
E_N &\leftrightarrow \text{ConL}(S_1, S_2, \ldots, S_N) \\
J_{ij} &\leftrightarrow -\frac{1}{\lambda} \log(1 - s_i s_j) \delta_{|A_i \cap A_j|}
\end{aligned}
\tag{45}
$$

Compared to the previous relations, the only change is the introduction of $\lambda$.

Considering the isomorphism between the Potts model and the graph model, the point of introducing $\lambda$ becomes clear. The bond probabilities, defined as $p_{ij} = 1 - e^{-\beta J_{ij}}$, can be written $J_{ij} = -1/\beta \log(1 - p_{ij})$, which show a great deal of resemblance to the pairwise weight of conflict. The inverse temperature $\beta$ may be identified with $\lambda$, and $\lambda$ may therefore get a physical interpretation. Moreover, the bond probabilities, $p_{ij}$, may be identified with the product of the support for the events $A_i$ and $A_j$ ($s_i s_j$). If there is no conflict between $A_i$ and $A_j$ ($A_i \cap A_j \neq \varnothing$), there is no bond in the corresponding graph model either. The isomorphisms between the three models are summarized as:

$$
\begin{aligned}
E_N &\leftrightarrow \text{ConL}(S_1, S_2, \ldots, S_N) \\
J_{ij} &\leftrightarrow -\frac{1}{\lambda} \log(1 - s_i s_j) \delta_{|A_i \cap A_j|} \\
\frac{1}{T} &= \beta \leftrightarrow \lambda \\
p_{ij} &\leftrightarrow s_i s_j, \quad \text{if conflict}, A_i \cap A_j = \varnothing
\end{aligned}
\tag{46}
$$

Thus, Dempster–Shafer clustering (if linearized) may be interpreted in terms of a graph model, where edges in the graph correspond to conflicting pieces of evidence, and the bond probability of an edge $p_{ij}$ is precisely the product of the support, $s_i s_j$.

Since $\beta$ plays a crucial role here, it is worth clarifying a few points. In statistical mechanics, the probability for a certain configuration with an energy $E_i$ is given by the Boltzmann distribution $e^{-\beta E_i}$. As the system freezes, $\beta \to \infty$, and only the state (or states) with the lowest energy gets an appreciable probability. Another way to view this is to consider the free energy $F$, defined as $F = E - TS$, where $S$ is the entropy of the system. A maximum likelihood estimate (which is equivalent with the mean field approximation) is obtained by minimizing $F$ subject to the constraint that the entropy is constant. The $\beta$ becomes a Lagrangian parameter that controls the average energy, and it gives the relative importance of energy minimization and entropy (disorder) maximization. As $T \to 0$, $\min F \Leftrightarrow \min E$, and it is expected that the absolute energy minimum state should be obtained as the ground state of the system.

It is important to note that in our analysis so far, $J_{ij}$'s are the independent parameters, $p_{ij}$'s the dependent ones, and $\beta$ parameterizes their relation. But for any given $\beta$ one could choose either the $J_{ij}$'s or the $p_{ij}$'s (or $s_i$'s) as the independent parameters.

However, if the $p_{ij}$'s are chosen as independent parameters, some of the previous discussion regarding the system's behavior during annealing may not be valid.

## 8
## Results

In this chapter we compare the clustering performance and computational complexity of three different methods for Dempster–Shafer clustering. First, the Potts spin clustering using simulated annealing discussed previously in this article. Secondly, the neural clustering inspired by the Hopfield and Tank [22] developed in [38] and further extended in [36, 37], is investigated, and finally, the iterative optimization that was initially developed for Dempster–Shafer clustering [29–35].

For each method, and all problem sizes, we will cluster $2^K - 1$ pieces of evidence into $K$ subsets. As reported before, the evidence supports all subsets of the frame $\Theta = \{1, 2, 3, \ldots, K\}$. Thus, there always exists a global minimum to the metaconflict function equal to zero, since all pieces of evidence that include the 1-element can be put into cluster 1, of the remaining evidence, all those that include the 2-element can be put into cluster 2, and so forth. Since all evidence of cluster 1 includes the 1-element their intersection is nonempty, and since all evidence of cluster 2 includes the 2-element their intersection is also nonempty, etc. Thus, all conflicts $c_i$ are zero and we always have a global minimum with $Mcf = 0$. It is easy to see that there are more than one completely conflict free solution of our benchmark problem.

The reason we choose a problem where the minimum metaconflict is zero is that it makes a good test example for evaluating performance. We have no reason to believe that this choice of test examples is atypical with respect to performance.

In Table 1 we notice that both neural methods have an exponential computation time in the number of items of clusters. This is solely due to the exponential growth in the number of items of evidence via $N = 2^K - 1$. However, the Potts spin clustering method is much faster than the neural clustering method inspired by Hopfield and Tank. For the problem of clustering 127 pieces of evidence into seven clusters, the Potts spin methods computation time is 8.78 s, while the Hopfield and Tank type of neural clustering had a computation time of 618 s, a difference of 70 times for this particular problem size. Although $K (= |\Theta|$; the number of clusters) and $N (= |2^\Theta| - 1$; the number of items of evidence) are not changed independently in these test examples, evidence is rather striking (Table 1) that the Potts spin computation time scales as $N^2 \log^2 N$. A small overhead is noted for the smallest problem sizes. The time complexity of iterative optimization is much worse than for both neural methods (see Fig. 5).

Let us also compare the clustering performance of the three methods. The cluster conflicts can vary strongly from cluster to cluster and occasionally larger conflicts are found. In Table 2 the median and mean metaconflict over ten runs is tabulated for each problem size and method. It is important to note that, although we are optimizing the linearized weight of conflict (Eq. (20)), it is the actual conflicts for each cluster that are calculated in Table 2. The Potts spin method is able to find a global optimum for problem sizes up to nine cluster. However, for the ten- and eleven-cluster problems the metaconflict increases rapidly, see Fig. 6. The same type of behavior is found for the Hopfield and Tank method but here the increase in metaconflict arises already with the six- and seven-cluster problems. Iterative optimization has a good performance for small problem sizes.

**Table 1.** Each number is the mean of ten randomly generated problems. For the Potts spin method, the total number of iterations needed depends strongly on the anneal parameters chosen. These parameters can be chosen to reduce the number of iterations without any significant performance decrease. All times are measured in seconds on an SGI Onyx computer (150 MHz CPU model IP19)

| No. of clusters, $K$ | No. of items of evidence, $N$ | Time | Potts spins | | Hopfield and Tank | | Iterative optimization | |
| --- | --- | --- | --- | --- | --- | --- | --- | --- |
| | | | Time/$N^2 K^2$ | Iterations | Time | Iterations | Time | Iterations |
| 3 | 7 | 0.026 | $5.90 \cdot 10^{-5}$ | 159 | 2.18 | 54.3 | 0.061 | 2.6 |
| 4 | 15 | 0.099 | $2.75 \cdot 10^{-5}$ | 244 | 7.75 | 63.8 | 0.201 | 5.1 |
| 5 | 31 | 0.455 | $1.89 \cdot 10^{-5}$ | 322 | 30.4 | 65.2 | 1.90 | 11.1 |
| 6 | 63 | 1.90 | $1.33 \cdot 10^{-5}$ | 367 | 109 | 79.8 | 288 | 26.1 |
| 7 | 127 | 8.78 | $1.11 \cdot 10^{-5}$ | 421 | 618 | 108 | 76* | |
| 8 | 255 | 46.8 | $1.12 \cdot 10^{-5}$ | 483 | | | | |
| 9 | 511 | 254 | $1.20 \cdot 10^{-5}$ | 507 | | | | |
| 10 | 1023 | 1290 | $1.23 \cdot 10^{-5}$ | 515 | | | | |
| 11 | 2047 | 6630 | $1.31 \cdot 10^{-5}$ | 534 | | | | |

*Days estimated

224

While Fig. 6 gives a good overview on the clustering performance of the three different methods, the picture might be somewhat misleading when evaluating performance. A large part of the increase in metaconflict is due to the increase in problem size. Each cluster contributes to the total metaconflict, and as the number of cluster increases, the total metaconflict increases as well. In order to eliminate this effect we must calculate the average metaconflict per cluster.

If we make an assumption that in a local minimum the conflict of each cluster is identical – which is certainly true for a global minimum – the average metaconflict per cluster may be calculated. With $c_0 = 0$ and $c_i = c_j$ for all $i, j$ we get from Eq. (14)

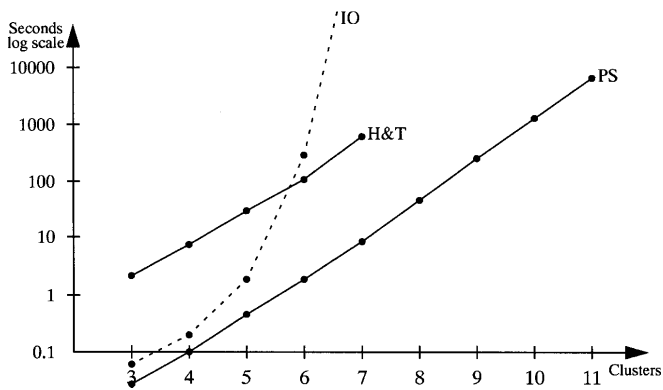

**Fig. 5.** Computation time (mean of 10 runs) for the three different methods; Iterative optimization (IO), Hopfield and Tank neural clustering (H&T), and Potts spin clustering (PS)

$$\langle c_i \rangle = 1 - (1 - Mcf)^{1/q} \quad . \qquad (47)$$

The median and mean metaconflicts per cluster are tabulated for all problem sizes in Table 3. Notice the smaller rise in metaconflict per cluster. For instance, the clustering performance of Potts spin for the ten-cluster problem is now visibly much better than that of the Hopfield and Tank for the seven-cluster problem, Fig. 7. In Fig. 7 we also observe the wide difference between median and mean metaconflict per cluster for the Potts spin method in the eleven-cluster problem. Hence, fluctuation of the results is large. While the method is still able to find some near optimal partitionings the average partition yields a higher metaconflict per cluster. This is a clear indication that the Potts method has reached its limit to produce perfect solutions, but still produces near optimal solutions. For slightly larger problem sizes we would still expect the method to find some reasonably good, but no longer near optimal partitions.

The best measure of clustering performance is the metaconflict per evidence. Simply divide the average metaconflict per cluster already found (in Table 3) with the average number of pieces of evidence in each cluster, see Table 4. This way we also take into account the exponential growth in the number of items of evidence as the number of clusters grow. The remarkable result is that the Potts model does not give any significant rise of the mean metaconflict per evidence, Fig. 8. This is true for almost three orders of magnitude.

For the eleven-cluster problem the Potts method achieves mean and median metaconflicts per evidence of just 0.8 and 2.4‰, respectively. We compare this with the

**Table 2.** Metaconflict (median and mean over ten runs) for the Potts spins, Hopfield and Tank and iterative optimization clustering methods

| No. of clusters, $K$ | No. of items of evidence, $N$ | Potts spins | | Hopfield and Tank | | Iterative optimization | |
|---|---|---|---|---|---|---|---|
| | | Median | Mean | Median | Mean | Median | Mean |
| 3 | 7 | 0 | 0 | 0.005 | 0.016 | 0 | 0 |
| 4 | 15 | 0 | 0 | 0.013 | 0.059 | 0 | 0.001 |
| 5 | 31 | 0 | 0 | 0.042 | 0.076 | 0 | 0.003 |
| 6 | 63 | 0 | 0.115 | 0.447 | 0.398 | 0 | 0.097 |
| 7 | 127 | 0 | 0.116 | 0.904 | 0.856 | | |
| 8 | 255 | 0 | 0.114 | | | | |
| 9 | 511 | 0.069 | 0.122 | | | | |
| 10 | 1023 | 0.711 | 0.610 | | | | |
| 11 | 2047 | 0.998 | 0.814 | | | | |

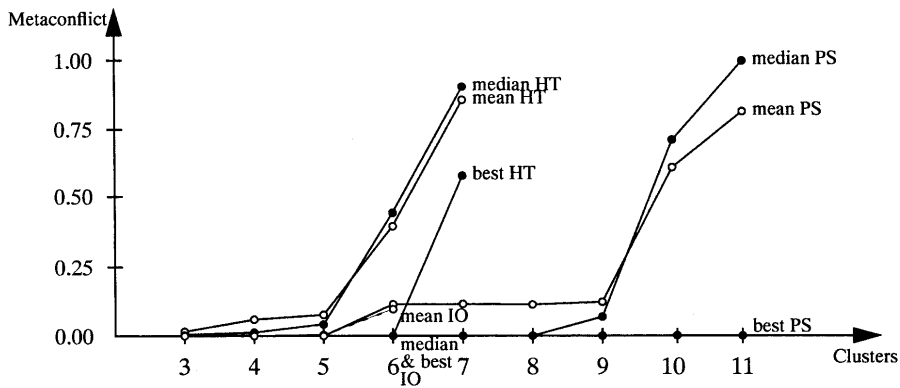

**Fig. 6.** Metaconflict

**Table 3.** Metaconflict per cluster, from Table 2 through $\langle c_i \rangle = 1 - (1 - Mcf)^{1/q}$

| No. of clusters, $K$ | No. of items of evidence, $N$ | Potts spins | | Hopfield and Tank | | Iterative optimization | |
| --- | --- | --- | --- | --- | --- | --- | --- |
| | | Median | Mean | Median | Mean | Median | Mean |
| 3 | 7 | 0 | 0 | 0.002 | 0.005 | 0 | 0 |
| 4 | 15 | 0 | 0 | 0.003 | 0.015 | 0 | 0.0003 |
| 5 | 31 | 0 | 0 | 0.009 | 0.016 | 0 | 0.0005 |
| 6 | 63 | 0 | 0.020 | 0.094 | 0.081 | 0 | 0.017 |
| 7 | 127 | 0 | 0.017 | 0.284 | 0.242 | | |
| 8 | 255 | 0 | 0.015 | | | | |
| 9 | 511 | 0.008 | 0.014 | | | | |
| 10 | 1023 | 0.117 | 0.090 | | | | |
| 11 | 2047 | 0.441 | 0.142 | | | | |

226

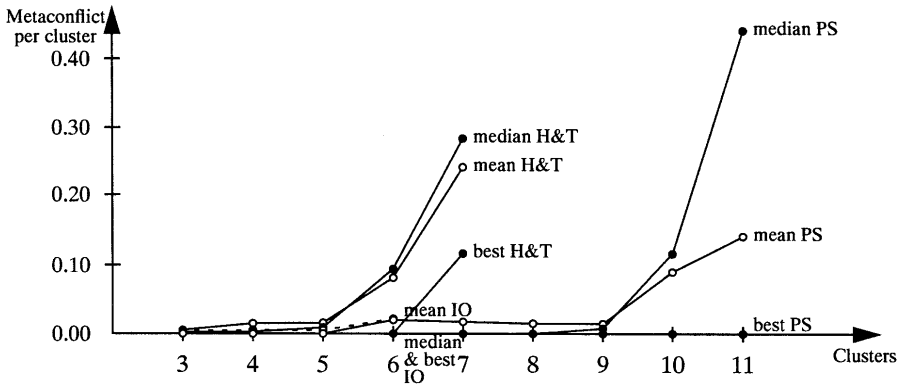

**Fig. 7.** Metaconflict per cluster, through $\langle c_i \rangle = 1 - (1 - Mcf)^{1/q}$

**Table 4.** Metaconflict per evidence

| No. of clusters, $K$ | No. of items of evidence, $N$ | Potts spins | | Hopfield and Tank | | Iterative optimization | |
| --- | --- | --- | --- | --- | --- | --- | --- |
| | | Median | Mean | Median | Mean | Median | Mean |
| 3 | 7 | 0 | 0 | 0.0007 | 0.002 | 0 | 0 |
| 4 | 15 | 0 | 0 | 0.0009 | 0.004 | 0 | 0.00009 |
| 5 | 31 | 0 | 0 | 0.001 | 0.003 | 0 | 0.00008 |
| 6 | 63 | 0 | 0.002 | 0.009 | 0.008 | 0 | 0.002 |
| 7 | 127 | 0 | 0.001 | 0.016 | 0.013 | | |
| 8 | 255 | 0 | 0.0005 | | | | |
| 9 | 511 | 0.0001 | 0.0003 | | | | |
| 10 | 1023 | 0.0011 | 0.0009 | | | | |
| 11 | 2047 | 0.0024 | 0.0008 | | | | |

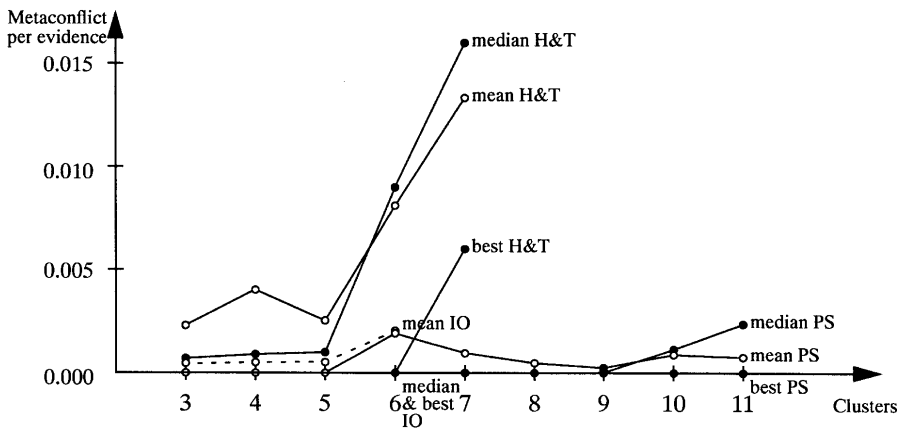

**Fig. 8.** Metaconflict per evidence

expected conflict between two random pieces of evidences, derived analytically.

With $2^q - 1$ pieces of evidence, all simple support functions with elements from the set of all subsets $2^\Theta$ of $\Theta = \{1, 2, 3, \ldots, q\}$, there are

$$\frac{1}{2}\left[(2^q - 1)^2 - (2^q - 1)\right] \tag{48}$$

possible pairs of two pieces of evidence. Of these,

$$\frac{1}{2}\sum_{j=1}^{n-1}\binom{n}{j}\sum_{k=1}^{n-j}\binom{n-j}{k} \tag{49}$$

are in conflict.

Thus, if two different pieces of evidence are drawn randomly from the set of all subsets, we have a probability of conflict between their propositions of

$$P(A_i \cap A_j = \varnothing) = \frac{\sum_{j=1}^{q-1}\binom{q}{j}\sum_{k=1}^{q-j}\binom{q-j}{k}}{(2^q - 1)^2 - (2^q - 1)} \ , \tag{50}$$

where, e.g., $P(A_i \cap A_j = \varnothing) = 4.13\%$ when $q = 11$.

The basic probability number of a piece of evidence is a uniformly distributed random number between 0 and 1. Thus, the expected conflict is 0.25 between two pieces of evidence that are known to be in conflict. Therefore, the expected conflict between two pieces of evidence drawn randomly from $2^\Theta$ becomes 0.0103 ($= 0.0413 \cdot 0.25$), which is roughly 4 times higher than the median, and 13 times larger than the mean metaconflict per evidence received in the eleven-cluster problem by the Potts spin method, Table 4. This is proof that on average the metaconflict per evidence obtained numerically corresponds to much less than one conflicting pair of pieces of evidence per cluster. It must be considered to be a very good result given that on average there are 186 pieces of evidence per cluster in the eleven-cluster problem, and had 186 pieces of evidence been drawn randomly from $2^\Theta$, then 710 pairs of the 17205 pairs in the cluster would be in conflict.

One final investigation is to see how often the three different clustering methods are able to find a global minimum of the metaconflict function over ten different runs, and with different random basic probability numbers assigned to the evidence for each run. Once again, we find, in Fig. 9, the superiority of the Potts spin method over the other two methods. It is able to find at least one global optimum with zero conflict over the ten runs for all problem sizes up to eleven clusters, except for the case with ten clusters where it got a very near miss with an overall metaconflict of 0.0013 for the best run. Such a miss is the result of one piece of evidence with a very small basic probability number being misplaced, yielding a very small conflict.

## 9
## Conclusions

There are two threads followed in this paper. First, for the pragmatic reader we have shown how the Potts model finds optimal or nearly optimal solutions to Dempster–Shafer clustering problems in every case. Additionally, the computation time scales as $N^2 \log^2 N$, when measured in the number of items of evidence. This is the same computational growth as in the previous Hopfield and Tank model, but with a much lower constant. Furthermore, the mean conflict per cluster and evidence shows no significant increase as the problem is scaled up. This is valid at least for problem sizes varying over three orders of magnitude.

Secondly, mainly of theoretical interest are the relations between the three models: linearized Dempster–Shafer, antiferromagnetic Potts spin model, and a graph optimization problem. The possibility to interpret the Dempster–Shafer clustering problem as a graph optimization problem is appealing.

Having developed a Potts spin model for Dempster–Shafer clustering with a near perfect clustering performance and a reasonably good computational performance for most problem sizes, we think this method can serve as a solution for preprocessing of data in many different applications, and especially so in the intended application for preprocessing of intelligence data in information fusion.

## References

1. **Bengtsson M** (1996) Hierarchical clustering in a mean field theory, FOA-R--96-00298-3.4--SE, Defence Research Establishment, Linköping
2. **Bengtsson M** (1998) Toy physics – Potts spins in clustering, FOA-R--98-00702-616--SE, Defence Research Establishment, Linköping
3. **Bengtsson M** (1998) Graph optimization and the Potts spin model, FOA-R--98-00860-616--SE, Defence Research Establishment, Linköping
4. **Bengtsson M** (1998) Combined clustering and topological ordering in a Potts model – numerical benchmarking studies, FOA-R--98-00930-616--SE, Defence Research Establishment, Linköping
5. **Bengtsson M** (1998) Nonparametric clustering in a mean field Potts model, FOA-R--98-00953-616--SE, Defence Research Establishment, Linköping
6. **Bengtsson M, Roivainen P** (1995) Using the Potts glass for solving the clustering problem, Int J Neural Syst **6**: 119–132
7. **Bergsten U, Schubert J** (1993) Dempster's rule for evidence ordered in a complete directed acyclic graph, Int J Approx Reason **9**: 37–73
8. **Bergsten U, Schubert J, Svensson P** (1997) Applying data mining and machine learning techniques to submarine intelligence analysis. In: Heckerman D, Mannila H, Pregibon D, Uthurusamy R (eds), Proc Third Int Conf Knowledge Dis-

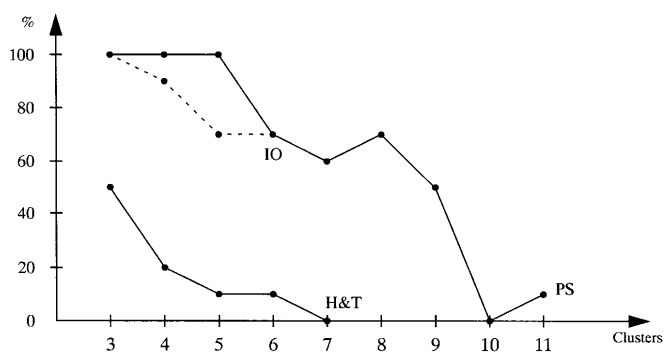

**Fig. 9.** Global optimum found in percent of runs

covery and Data Mining (KDD'97), pp. 127–130, Newport Beach, AAAI Press, Menlo Park, USA

9. **Blatt M, Wiseman S, Domany E** (1996) Clustering data through an analogy to the potts model. In: Touretzky DS, Mozer MC, Hasselmo ME (eds), Advances in Neural Information Processing Systems 8, Proc Conf Neural Information Processing Systems 1995 (NIPS'95), pp. 416–422, MIT Press, Denver, Cambridge, USA

10. **Blatt M, Wiseman S, Domany E** (1996) Super-paramagnetic clustering of data, Phys Rev Lett **76**: 3251–3255

11. **Blatt M, Wiseman S, Domany E** (1997) Data clustering using a model granular magnet, Neural Comput **9**: 1805–1842

12. **Buhmann J, Kühnel H** (1993) Complexity optimized data clustering by competitive neural networks, Neural Comput **5**: 75–88

13. **Buhmann J, Kühnel H** (1993) Vector quantization with complexity cost, IEEE Trans Info Theory **39**: 1133–1145

14. **Dempster AP** (1967) Upper and lower probabilities induced by a multiple valued mapping, Ann Math Statist **38**: 325–339

15. **Dempster AP** (1968) A generalization of Bayesian inference, J R Statist Soc B **30**: 205–247

16. **Dotsenko V** (1994) An introduction to the theory of spin glasses and neural networks, World Scientific Publishers, Singapore

17. **Edwards RG, Sokal AD** (1988) Generalization of the Fortuin–Kastelyn–Swendsen–Wang representation and Monte Carlo algorithm, Phys Rev D **38**: 2009–2012

18. **Fortuin CM, Kastelyn PW** (1972) On the random-cluster model, Physica **57**: 536–564

19. **Gislén L, Peterson C, Söderberg B** (1989) Teachers and classes with neural networks, Int J Neural Syst **1**: 167–176

20. **Gislén L, Peterson C, Söderberg B** (1992) Complex scheduling with Potts neural networks, Neural Comput **4**: 805–831

21. **Hopfield JJ** (1982) Neural networks and physical systems with emergent collective computational abilities, Proc Natl Acad Sci USA **79**: 2554–2558

22. **Hopfield JJ, Tank D** (1985) "Neural" computation of decisions in optimization problems, Biol Cybern **52**: 141–152

23. **Häkkinen J, Lagerholm M, Peterson C, Söderberg B** (1998) A Potts neuron approach to communication routing, Neural Comput **10**: 1587–1599

24. **Kanter I, Sompolinsky H** (1987) Graph optimization problems and the Potts glass, J Physics A **20**: L673–L679

25. **Ohlsson M, Peterson C, Yuille AL** (1992) Track finding with deformable templates – The elastic arms approach, Computer Physics Comm **71**: 77–98

26. **Peterson C, Anderson JA** (1988) Neural networks and NP-complete optimization problems: A performance study on the graph bisection problem, Complex Systems **2**: 59–89

27. **Peterson C, Söderberg B** (1989) A new method for mapping optimization problems onto neural networks, Int J Neural Syst **1**: 3–22

28. **Rose K, Gurewitz E, Fox G** (1990) Statistical mechanics and phase transitions in clustering, Phys Rev Lett **65**: 945–948

29. **Schubert J** (1993) On nonspecific evidence, Int J Intell Syst **8**: 711–725

30. **Schubert J** (1994) Cluster-based specification techniques in Dempster–Shafer theory for an evidential intelligence analysis of multiple target tracks, PhD Thesis, TRITA-NA-9410, ISRN KTH/NA/R--94/10--SE, ISSN 0348-2952, ISBN 91-7170-801-4, Royal Institute of Technology, Stockholm

31. **Schubert J** (1995) Cluster-based specification techniques in Dempster–Shafer theory for an evidential intelligence analysis of multiple target tracks (thesis abstract), AI Comm **8**: 107–110

32. **Schubert J** (1995) Finding a posterior domain probability distribution by specifying nonspecific evidence, Int J Uncertainty, Fuzziness and Knowledge-Based Syst **3**: 163–185

33. **Schubert J** (1995) Cluster-based specification techniques in Dempster–Shafer theory. In: Froidevaux C, Kohlas J (eds), Symbolic and quantitative approaches to reasoning and uncertainty, Proc European Conf Symbolic and Quantitative Approaches to Reasoning and Uncertainty (ECSQARU'95), pp. 395–404, Fribourg, Switzerland 1995, Springer-Verlag (LNAI 946), Berlin

34. **Schubert J** (1996) Specifying nonspecific evidence, Int J Intell Syst **11**: 525–563

35. **Schubert J** (1997) Creating prototypes for fast classification in Dempster–Shafer clustering. In: Gabbay DM, Kruse R, Nonnengart A, Ohlbach HJ (eds), Qualitative and quantitative practical reasoning, Proc First Int J Conf Qualitative and Quantitative Practical Reasoning (ECSQARU-FAPR'97), pp. 525–535, Bad Honnef, Germany 1997, Springer-Verlag (LNAI 1244), Berlin

36. **Schubert J** (1998) A neural network and iterative optimization hybrid for Dempster–Shafer clustering. In: Bedworth M, O'Brien J (eds), Proc EuroFusion98 Int Conf Data Fusion (EuroFusion'98), pp. 29–36, Great Malvern, UK

37. **Schubert J** (1999) Simultaneous Dempster–Shafer clustering and gradual determination of number of clusters using a neural network structure. In: Proc 1999 Information, Decision and Control Conf (IDC'99), pp. 401–406, Adelaide, Australia 1999, IEEE, Piscataway

38. **Schubert J** (1999) Fast Dempster–Shafer clustering using a neural network structure. In: Bouchon–Meunier B, Yager RR, Zadeh LA (eds), Information, uncertainty and fusion, pp. 419–430, Kluwer Academic Publishers (SECS 516), Boston

39. **Schubert J** (2000) Managing inconsistent intelligence. In: Proc Third Int Conf Information Fusion (FUSION 2000), pp. TuB4/10–16, Paris, France 2000, Int Soc Information Fusion, Sunnyvale

40. **Shafer G** (1976) A mathematical theory of evidence, Princeton University Press, Princeton

41. **Shafer G** (1990) Perspectives on the theory and practice of belief functions, Int J Approx Reason **4**: 323–362

42. **Shafer G** (1992) Rejoinder to comments on "Perspectives on the theory and practice of belief functions," Int J Approx Reason **6**: 445–480

43. **Smets P** (1999) Practical uses of belief functions. In: Laskey KB, Prade H (eds), Proc Fifteenth Conf Uncertainty in Artificial Intelligence (UAI'99), pp. 612–621, Stockholm, Sweden 1999, Morgan Kaufmann Publishers, San Francisco

44. **Smets P, Kennes R** (1994) The transferable belief model, Artif Intell **66**: 191–234

45. **Wu FY** (1983) The Potts Model, Rev Modern Physics **54**: 235–268

46. **Yager RR, Fedrizzi M, Kacprzyk J** (eds) (1994) Advances in the Dempster–Shafer theory of evidence, John Wiley & Sons, New York